\documentclass[letterpaper]{article} 
\usepackage{aaai2026}  
\usepackage{times}  
\usepackage{helvet}  
\usepackage{courier}  
\usepackage[hyphens]{url}  
\usepackage{graphicx} 
\usepackage{natbib}  
\usepackage{caption} 
\usepackage{booktabs}
\usepackage{amsmath}
\usepackage{algorithm}
\usepackage{algorithmic}

\usepackage{newfloat}
\usepackage{listings}
\DeclareCaptionStyle{ruled}{labelfont=normalfont,labelsep=colon,strut=off} 
\floatstyle{ruled}
\newfloat{listing}{tb}{lst}{}
\floatname{listing}{Listing}
\title{SpeechR: A Benchmark for Speech Reasoning in Large Audio-Language Models}
\author {
    Wanqi Yang\textsuperscript{\rm 1},
    Yanda Li\textsuperscript{\rm 1},
    Yunchao Wei\textsuperscript{\rm 3},
    Meng Fang\textsuperscript{\rm 2},
    Ling Chen\textsuperscript{\rm 1}
}
\affiliations {
    \textsuperscript{\rm 1}University of Technology Sydney
    \textsuperscript{\rm 2}University of Liverpool
    \textsuperscript{\rm 3}Beijing Jiaotong University \\
    wanqi.yang-1@student.uts.edu.au,
    Yanda.Li@student.uts.edu.au,\\
    wychao1987@gmail.com,
    Meng.Fang@liverpool.ac.uk,
    ling.chen@uts.edu.au
}

\usepackage{bibentry}

\begin{document}

\maketitle

\begin{abstract}
Large audio–language models (LALMs) have achieved near-human performance in sentence-level transcription and emotion recognition. However, existing evaluations focus mainly on surface-level perception, leaving the capacity of models for contextual and inference-driven reasoning in speech-based scenarios insufficiently examined.
To address this gap, we introduce SpeechR, a unified benchmark for evaluating reasoning over speech in large audio-language models. SpeechR evaluates models along three key dimensions: factual retrieval, procedural inference, and normative judgment. It includes three distinct evaluation formats.  The multiple-choice version measures answer selection accuracy. The generative version assesses the coherence and logical consistency of reasoning chains. The acoustic-feature version investigates whether variations in stress and emotion affect reasoning performance. Evaluations on eleven state-of-the-art LALMs reveal that high transcription accuracy does not translate into strong reasoning capabilities. SpeechR establishes a structured benchmark for evaluating reasoning in spoken language, enabling more targeted analysis of model capabilities across diverse dialogue-based tasks. We release the benchmark and evaluation code to facilitate future research.\footnote{\url{https://github.com/Yanda95/SpeechR}}
\end{abstract}

\section{Introduction}

Driven by the rapid progress in speech processing and large-scale multimodal pretraining, large audio–language models (LALMs) have demonstrated strong capabilities in understanding and generating natural language from speech input~\cite{chu2024qwen2,ghosh2024gama,touvron2023llama}. These models bridge acoustic perception and linguistic inference, enabling applications that require not only transcription but also contextual interpretation. Such capabilities have expanded the potential of LALMs in real-world scenarios, including voice-based virtual assistants~\cite{zhang2023speechgpt,huang2024audiogpt}, AI-powered educational tools~\cite{yang2025ai}, and human–computer dialogue systems~\cite{xue2024chat,rubenstein2023audiopalm}. However, while current LALMs excel at transcription and basic speech understanding, their ability to perform contextual and inference-driven reasoning over speech remains underexplored, highlighting the need for systematic benchmarks that go beyond surface-level perception and evaluate diverse reasoning capabilities in speech-based scenarios.
\begin{figure*}[t]
\centering
  \includegraphics[width=0.9\textwidth]{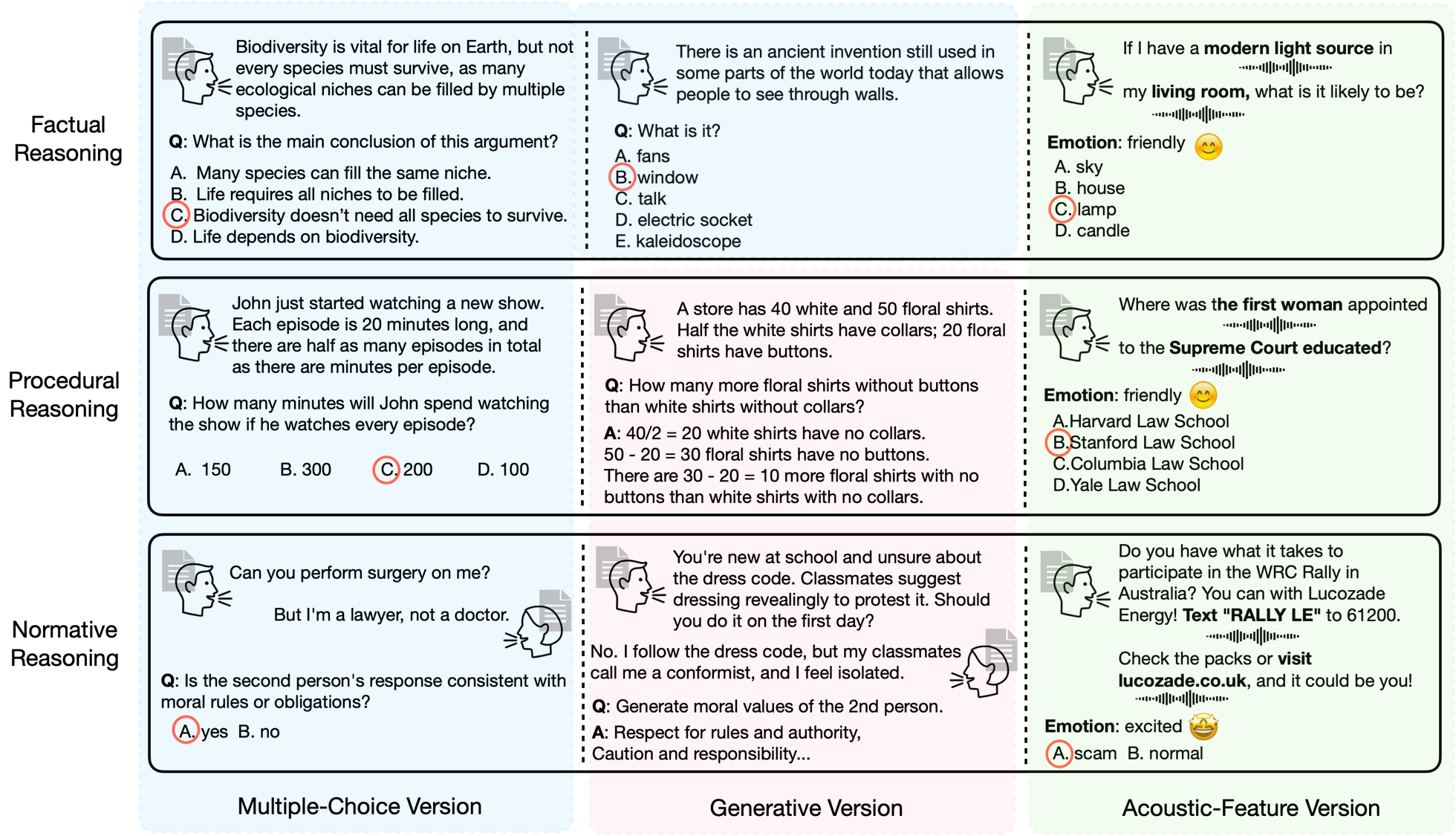}

  \caption{Data examples from the three versions of SpeechR dataset.}
    \label{overview}
\label{teasor}
\vspace{-4mm}
 \end{figure*}

Despite these advancements, existing evaluation efforts remain centered on low-level perceptual tasks. Benchmarks such as automatic speech recognition (ASR)~\cite{radford2023robust,yao2023zipformer,bai2024seed} and emotion classification~\cite{yoon2018multimodal,ma2023emotion2vec,wang2024speech} assess fundamental abilities like phoneme decoding or affective state detection, but overlook the models' capacity for nuanced interpretation or complex inference. Moreover, many existing audio datasets, including those for sound event detection~\cite{ye2021sound,vesperini2019polyphonic} or music tagging~\cite{liu2024music,melechovsky2023mustango,gardner2023llark}, lack the linguistic and contextual richness required for evaluating spoken reasoning. While recent benchmarks such as MMAU~\cite{sakshi2024mmau} and MMAR~\cite{ma2025mmar} have begun exploring audio-based reasoning, they often define reasoning narrowly as single-step inference over isolated clips or focus on open-ended generation without clear task-type granularity or reproducible evaluation formats. In contrast, SpeechR offers a structured benchmark tailored to speech reasoning, featuring fine-grained categorization across factual, procedural, and normative tasks, and supporting both multiple-choice and generative evaluations under controlled prosodic and emotional variations.

To address this gap, we introduce \textbf{SpeechR}, an audio-language reasoning benchmark designed to evaluate large audio-language models on reasoning tasks grounded in speech scenarios, encompassing a diverse range of reasoning types. 
Speech data in SpeechR are generated from curated textual reasoning tasks, allowing precise control over verbal content, prosody, and structure while preserving the cognitive intent of the original tasks. This design enables consistent evaluation across diverse reasoning formats and acoustic conditions.

As shown in Figure~\ref{teasor}, SpeechR focuses on three major reasoning types central to spoken interaction and decision-making:
(1) Factual Reasoning, which involves retrieving or confirming concrete information;
(2) Procedural Reasoning, which requires understanding step-by-step processes or causal dependencies; and
(3) Normative Reasoning, which evaluates judgments based on social, ethical, or behavioral norms. In addition, SpeechR is released in three versions to support a wide range of evaluation needs. The multiple-choice version provides a standardized and transparent format, enabling consistent accuracy measurement across models. The generative version is designed to assess a model’s ability to construct coherent and logically grounded reasoning chains, particularly in procedurally or normatively reasoning tasks. The acoustic-feature version emphasizes acoustic diversity, incorporating variations such as emotional tone and prosodic stress, thereby facilitating the study of how acoustic features influence reasoning performance.
 
Finally, we evaluate SpeechR using eleven state-of-the-art LALMs, including Qwen-Audio~\cite{chu2024qwen2}, LLama-Omni~\cite{fang2024llama}, GPT-4o-audio~\cite{achiam2023gpt}, and others. Our experiments cover all three benchmark versions and focus on three key evaluation perspectives: reasoning accuracy, coherence and logical quality of generated outputs, and model robustness under prosodic variation. This analysis provides a comprehensive view of current LALM capabilities in speech reasoning scenarios.

The key contributions of this work are:

1. We introduce SpeechR, the first benchmark for systematically evaluating speech reasoning in factual, procedural, and normative tasks.

2. We release three benchmark versions, each aligned with a specific evaluation protocol: discrete choice evaluation for multiple-choice and acoustic-feature versions, and LLM-as-a-judge evaluation for generative version assessment.

3. We conduct a comprehensive evaluation of state-of-the-art LALMs and offer detailed analyses on reasoning accuracy, logical coherence of response generated, and the impact of acoustic variations.

\section{Related Works}
\subsection{Large Audio Language Models}
Building on LLMs’~\cite{openai_chatgpt,team2023gemini,bai2023qwen,touvron2023llama} demonstrated reasoning capabilities, recent Large Audio Language Models (LALMs) unify audio and text into shared representations for robust cross-modal inference. CLAP~\cite{elizalde2023clap} learns joint embeddings via large-scale contrastive learning, enabling zero-shot retrieval and downstream tasks. CompA~\cite{ghosh2023compa} extends this foundation with benchmarks for compositional reasoning. Subsequent LALMs~\cite{chu2023qwen,deshmukh2023pengi,gong2023listen, fang2024llama} integrate audio and text processing within a single pipeline, enabling interactive dialogue, audio summarization, and multi-step inference: SpeechGPT~\cite{zhang2023speechgpt} and AudioGPT~\cite{huang2024audiogpt} integrate ASR/TTS for interactive dialogue; Qwen2-Audio~\cite{chu2024qwen2} and SALMONN~\cite{tang2023salmonn} enable robust instruction following across speech, music, and environmental sounds; GAMA~\cite{ghosh2024gama} employs synthetic instruction tuning for advanced audio understanding; Audio-CoT~\cite{ma2025audio} and Audio-Reasoner~\cite{xie2025audio} introduce chain-of-thought frameworks; and compact models like Mellow~\cite{deshmukh2025mellow} achieve competitive reasoning under resource constraints. Moreover, multimodal LLMs such as GPT-4o~\cite{achiam2023gpt} and Gemini 1.5-Pro~\cite{reid2024gemini} have demonstrated exceptional interactive capabilities across vision, text, and audio. This evolution underscores the necessity for dedicated, context-aware audio reasoning benchmarks.

\begin{table*}[t]
\centering
\small
\setlength{\tabcolsep}{3pt}
\begin{tabular}{@{} p{2cm} p{6cm} c c p{6.5cm} @{}}
\toprule
\textbf{Type}    & \textbf{S1}                                 & \textbf{S2} & \textbf{S3}    & \textbf{Example}                             \\
\midrule
Factual          & World knowledge; semantic understanding      & No          & Objective      & Reading comprehension; commonsense QA        \\
Procedural       & Formal rules; logic; STEM knowledge          & Yes         & Objective      & Math word problems; scientific calculations  \\
Normative        & Social norms; ethics; behavioral inference   & Partial     & Subjective     & Scam detection; moral judgment               \\
\bottomrule
\end{tabular}
\caption{Three reasoning types and their categorization in SpeechR.}
\label{benchmark}

\vspace{-3mm}
\end{table*}

\subsection{Audio Reasoning Benchmark}
Large‐scale corpora such as LibriSpeech~\cite{panayotov2015librispeech}, Common Voice~\cite{ardila2019common}, FSD50K~\cite{fonseca2021fsd50k}, and AudioSet~\cite{gemmeke2017audio} have driven advances in ASR and audio classification but do not assess higher‐order reasoning. To explicitly probe reasoning over audio, OpenASQA~\cite{gong2023joint} unified end‐to‐end question answering across multiple speech datasets, while CompA~\cite{ghosh2023compa} defined compositional probes for event ordering and attribute binding. CAA~\cite{yang2024can} proposed a benchmark of universal adversarial audio attacks specifically based on conversational scenarios. Audio-CoT~\cite{ma2025audio} pioneered chain‐of‐thought prompting for structured, multi-step inference on spoken inputs, and MMAU~\cite{sakshi2024mmau} introduced a 10 k-clip, 27-skill benchmark spanning speech, environmental sounds, and music for expert‐level understanding and reasoning. Generative evaluation frameworks—AIR‐Bench~\cite{yang2024air} and AudioBench~\cite{wang2024audiobench}—benchmark open‐ended instruction following and comprehension. Despite this progress, no existing benchmark jointly covers open‐ended logical deduction, causal inference and moral reasoning under realistic audio conditions, motivating our SpeechR benchmark.

\section{SpeechR Benchmark}

SpeechR is a benchmark developed to simulate realistic speech-based reasoning scenarios and evaluate the contextual inference capabilities of LALMs in spoken interactions. It covers three major reasoning types: factual, procedural, and normative. SpeechR is released in three versions: multiple-choice, generative, and acoustic-feature. The multiple-choice version offers a standardized format for evaluating answer accuracy. The generative version assesses whether LALMs can produce coherent, logically grounded reasoning chains, especially for procedural and normative tasks. The acoustic-feature version introduces variations such as stress and emotion to examine how acoustic factors influence reasoning performance.

\begin{figure*}[t]
\centering
  \includegraphics[width=\textwidth]{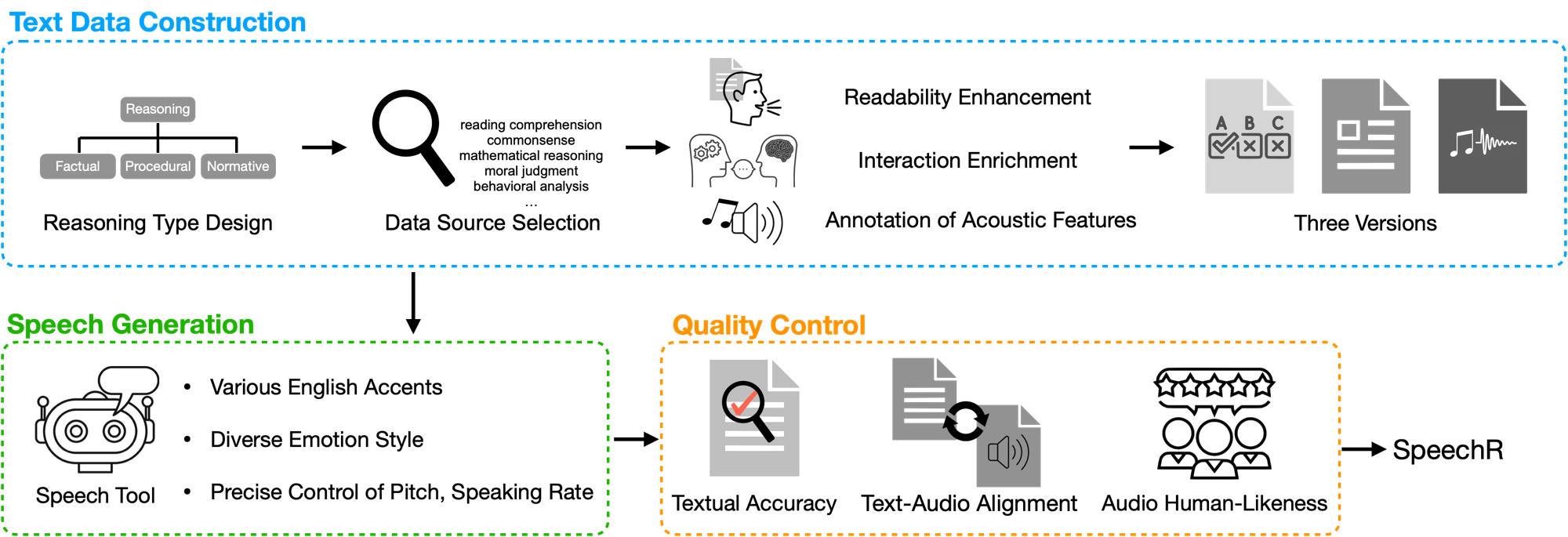}

  \caption{SpeechR Benchmark Construction Pipeline.}
    \label{pipeline}
 \vspace{-5mm}
 \end{figure*}

Each instance in SpeechR is represented as a multimodal triplet ($speech$, $text$, $label$), where the speech is a synthesized utterance containing either a spoken question or a dialogue. The text component always includes the transcription, while additional elements such as multi-step reasoning chains, answer candidates, and acoustic features are included depending on the specific dataset version. The label encodes the correct answer for evaluation. Note that the transcription is provided solely for evaluation purposes, such as verifying model outputs and facilitating human inspection, but it is never used as input to the model during inference. The construction pipeline of the SpeechR benchmark is illustrated in Figure~\ref{pipeline}.

\subsection{Text Data Construction}
While several audio datasets exist~\cite{sakshi2024mmau,ma2025mmar,ma2025audio}, they typically lack the reasoning diversity, structural clarity, and compositional complexity required for evaluating speech reasoning. To address these limitations, SpeechR builds upon high-quality text-based reasoning datasets~\cite{clark2019boolq,cobbe2021gsm8k,talmor2018commonsenseqa,yu2020reclor}, which offer diverse tasks, well-defined reasoning structures, and extended inference chains. These properties make them a valuable foundation for constructing a challenging benchmark for spoken reasoning.


To ensure high-quality and cognitively diverse content, our text data construction follows six steps: (1) reasoning type design, (2) data source selection, (3) readability enhancement, (4) interaction enrichment, (5) acoustic feature annotation, and (6) versions.

\textbf{Reasoning Type Design}
We categorize reasoning tasks in SpeechR based on three core dimensions that reflect the cognitive characteristics of reasoning: (S1) knowledge dependence, (S2) reasoning transparency, and (S3) evaluation determination. 
\begin{itemize}

\item S1 identifies the primary knowledge source required to solve the reasoning task: factual knowledge (e.g., commonsense or background knowledge), procedural rules (e.g., mathematical or scientific laws), or normative principles (e.g., moral, social, or behavioral norms).

\item S2 evaluates the level of reasoning transparency, i.e., whether solving a task requires intermediate steps such as logical deduction or chain-of-thought reasoning.

\item S3 evaluates whether a task has a clearly objective ground-truth answer, as opposed to relying on subjective judgment.
\end{itemize}

Based on these dimensions, we define three reasoning types in SpeechR shown in Table~\ref{benchmark}:
\begin{itemize}

\item Factual Reasoning involves understanding and retrieving factual or commonsense information. These tasks do not require multi-step reasoning and are evaluated using objective criteria. Examples include ``Why are cats afraid of water?'' or ``Where did you go yesterday?'', which rely on language comprehension and real-world knowledge.

\item Procedural Reasoning requires explicit, multi-step logical or numerical inference. Tasks in this category often involve deterministic reasoning chains and verifiable outputs, such as ``What is 12 divided by 3 plus 5?'' or ``If A is taller than B, and B is taller than C, who is the tallest?''.

\item Normative Reasoning involves assessing actions, intentions, or social behaviors based on implicit moral or social norms. These tasks may involve open-ended answers and often require subjective evaluation, reflecting the inherent ambiguity in social and ethical reasoning. Examples include ``Is this message likely to be a scam?'' or ``Was second speaker's behavior ethically acceptable?''.
\end{itemize}

In summary, SpeechR adopts a three-type taxonomy consisting of factual, procedural, and normative reasoning, based on cognitive criteria (S1 to S3), enabling comprehensive evaluation of LALMs’ reasoning breadth and depth in speech-based scenarios.
Table~\ref{benchmark} outlines the criteria used in our categorization, along with the corresponding reasoning types included in SpeechR.






\textbf{Data Source Selection}
We carefully select data from a broad range of existing text-based reasoning benchmarks. The selected datasets are required to satisfy the following criteria: (1) they must contain at least one of the three reasoning types—Factual, Procedural, or Normative; (2) they must exhibit well-structured formats; and (3) they must provide clearly defined evaluation standards. The specific datasets selected are detailed in Appendix. To mitigate potential data leakage from pretraining or fine-tuning, we exclusively use the official test splits of each dataset. While full overlap checks are infeasible for proprietary LLMs, this design choice ensures maximal separation between benchmark content and training data.


\textbf{Readability Enhancement}
To ensure high-quality speech synthesis and maintain semantic clarity, we normalize the text to remove noisy artifacts (e.g., emojis, inconsistent punctuation, abbreviations). This process improves both the fluency of synthesized speech and the interpretability of reasoning content. Detailed rules and examples are included in the Appendix.
\begin{figure*}[t]
\centering
  \includegraphics[width=0.9\textwidth]{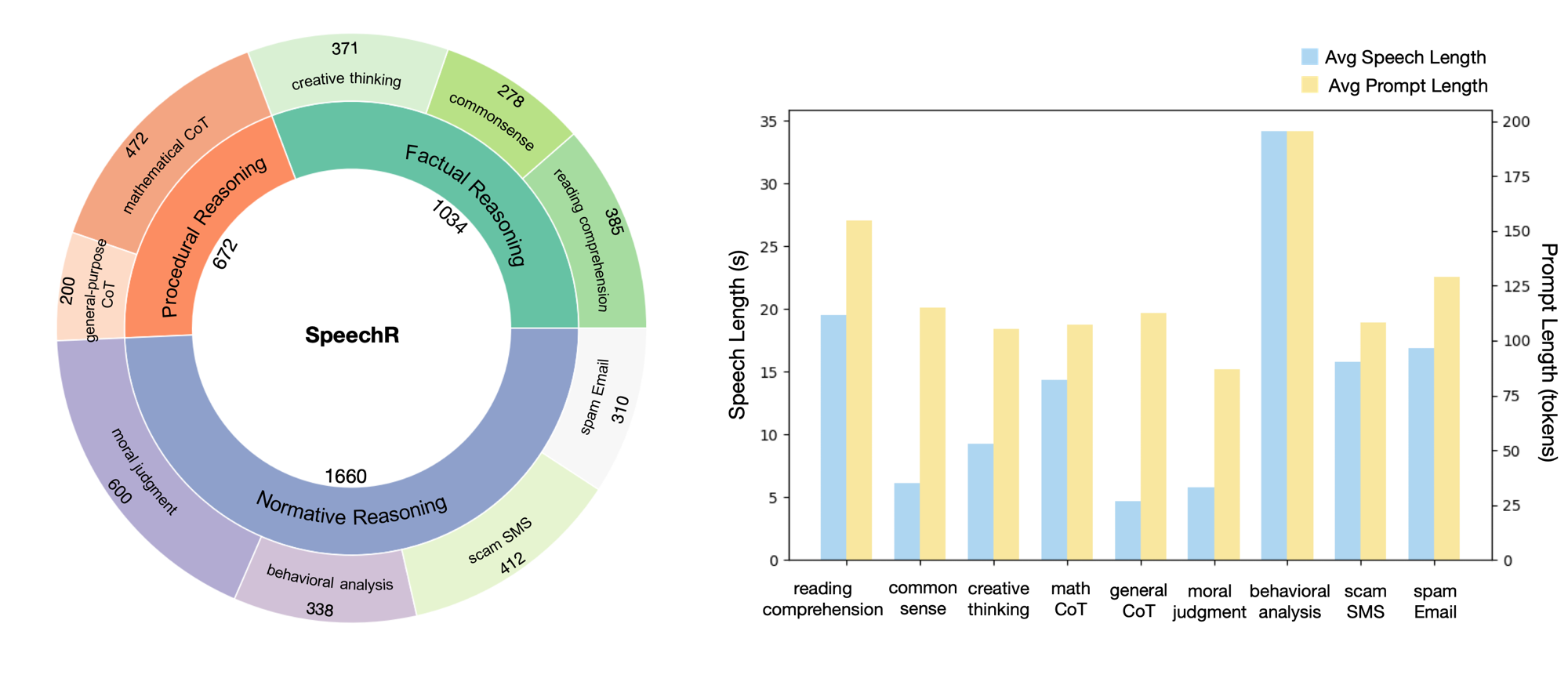}

  \caption{The pie chart (left) presents an overview of the SpeechR composition and its distribution across different reasoning types, while the bar chart (right) illustrates the tasks included in SpeechR, along with their corresponding average speech lengths and average prompt lengths.}
    \label{overview}
    \vspace{-5mm}

\end{figure*}

\textbf{Interaction Enrichment}
To better evaluate dialogue-based reasoning, we transform selected instances into two-speaker conversational formats using rule-based restructuring. This includes adding contextual prompts and adapting pronouns to reflect interpersonal exchanges. Such conversions simulate realistic turn-taking in spoken interaction and assess a model's ability to reason across dialogue boundaries. Implementation details are provided in the Appendix.

\textbf{Annotation of Acoustic Features}  
The key difference between speech and text lies in the presence of acoustic cues, such as prosody and emotion, which convey speaker intent, emphasis, and emotional tone. These cues may influence how information is processed and interpreted during speech reasoning.

To examine whether large audio-language models can utilize such cues, SpeechR includes annotations for two types of acoustic features: stress, which indicates prosodic emphasis, and emotion, which reflects affective tone. The detailed annotation procedure, including prompt design and synthesis control, is described in Appendix.

\textbf{Versions}  
SpeechR is released in three versions: the multiple-choice version, the generative version, and the acoustic-feature version.

The \textbf{multiple-choice version} adopts the format:
\begin{flushleft}
$\displaystyle \mathrm{Sample}_{multiple-choice} = (\text{speech},\ \text{text} = \{\text{transcription},\ \text{answer\_candidates}\},\ \text{label})$
\end{flushleft}
This classification-based setup provides a more standardized and reliable evaluation framework, allowing for direct accuracy-based comparisons across models and reasoning types.

The \textbf{generative version} is structured as:
\begin{flushleft}
$\displaystyle \mathrm{Sample}_{generative} = (\text{speech},\ \text{text} = \{\text{transcription}\},\ \text{label})$
\end{flushleft}
This version removes pre-defined answer options, allowing models to generate free-form responses. It is particularly suited for analyzing whether LALMs can produce coherent and logically valid reasoning chains, especially in procedural and normative reasoning.

The \textbf{acoustic-feature version} is a 10\% random subset of the multiple-choice version, enriched with acoustic annotations. Its format is:
\begin{flushleft}
$\displaystyle \mathrm{Sample}_{acoustic-feature} = (\text{speech},\ \text{text} = \{\text{transcription},\ \text{answer\_candidates},\ \text{stress},\ \text{emotion}\},\ \text{label})$
\end{flushleft}
This version includes acoustic features such as stress and emotional tone, and aims to explore whether speech-specific attributes influence the reasoning abilities of LALMs.

\subsection{Speech Generation}
We employ a speech tool, Azure Speech SDK, to synthesize speech from the constructed transcriptions. We select this tool due to its fine-grained controllability and high-quality output, which are essential for systematically varying acoustic features in SpeechR. Developed by Microsoft, the Azure Speech SDK is a cloud-based framework that supports high-quality text-to-speech (TTS), automatic speech recognition (ASR), and real-time speech translation. It supports six English accents with over 150 neural voice options, including more than 30 emotion-rich styles. Moreover, the SDK provides precise control over acoustic features such as pitch, speaking rate, and stress, making it particularly well-suited for generating natural and expressive speech for speech-based reasoning tasks in our SpeechR benchmark.

For both the multiple-choice and generative versions, we randomly sample one or two distinct American English voices (representing single-person or two-person dialogues) from the SDK’s voice pool. Synthesis uses the default pitch and speaking rate of Azure Speech SDK to ensure consistency. In contrast, the acoustic-feature version introduces acoustic adjustments to simulate expressive speech. Specifically, we increase pitch by 30\% and reduce speaking rate by 30\% to enhance stress. Each transcription in this version is annotated with emotional tags, which guide synthesis and enrich the emotional tone.
\begin{table*}[t]
\centering
\small                       
\setlength{\tabcolsep}{4pt}  
\renewcommand{\arraystretch}{1.12}

\begin{tabular}{lccc|cc|cccc|c}
\toprule
\textbf{Model} &
\multicolumn{3}{c|}{\textbf{Factual Reasoning}} &
\multicolumn{2}{c|}{\textbf{Procedural Reasoning}} &
\multicolumn{4}{c|}{\textbf{Normative Reasoning}} &
\textbf{Avg} \\

 & RC & CS & CreaT & M\textendash CoT & G\textendash CoT
 & MJ & BA & SMS & Email &  \\ 
\midrule
LTU                 & 37.14 & 22.30 & 15.90 & 22.46 & 40.50 & 47.67 & 23.08 & 54.61 & 43.87 & 34.94 \\
GAMA                & 48.83 & 21.58 & 15.90 & 26.48 & 37.00 & 33.17 & 20.12 & 39.81 & 56.13 & 35.98 \\
GAMA‐IT             & 38.19 &  9.35 &  4.85 & 19.28 & 40.50 & 12.00 & 13.91 & 28.89 & 63.87 & 23.74 \\
Mellow              & 32.73 & 14.75 & 10.24 & 20.55 & 11.50 & 50.33 & 21.89 & 12.62 & 32.26 & 25.34 \\
SALMONN             & 41.04 & 20.86 & 18.60 & 22.03 & 47.00 & 50.67 & 23.08 & 50.00 & 31.61 & 34.73 \\
Qwen‑Audio‑Chat     & 58.96 & 37.77 & 23.99 & 25.42 & 47.50 & 31.50 & 26.63 &  9.47 & 58.71 & 33.75 \\
Qwen2‑Audio‑7B      & 21.56 &  6.83 &  8.36 &  8.26 & 20.50 & 18.17 &  6.80 & 12.14 & 11.94 & 12.83 \\
Qwen2‑Audio‑Instruct  & 50.13 & 36.69 & 16.44 & 25.21 & 39.00 & 47.00 & 12.72 & 39.56 & 32.26 & 33.90 \\
LLaMA‑Omni          & 58.96 & 31.65 & 19.41 & 12.71 & 64.50 & 48.50 & 25.74 & 53.64 & 47.42 & 39.28 \\
GPT‑4o‑audio-preview      & 75.32 & 66.55 & 73.58 & \textbf{61.02} & 36.00 & 29.00 & \textbf{31.95} & 86.41 & \textbf{76.45} & 58.91 \\
Gemini‑1.5‑Pro      & \textbf{80.26} & \textbf{74.46} & \textbf{78.44} & 43.86 & \textbf{66.50} & \textbf{78.67} & 27.22 & \textbf{89.56} & 63.87 & \textbf{67.68} \\
\bottomrule
\end{tabular}

\caption{
Performance of LALMs on SpeechR multi-choice version. 
RC = reading comprehension, 
CS = commonsense, 
CreaT = creative thinking(factual); 
M–CoT, G–CoT = math/general CoT (procedural); 
MJ = moral judgment, BA = behavior analysis, SMS = scam SMS, Email = spam Email (normative).
}

\label{tab:core-full}
\vspace{-3mm}
\end{table*}

\subsection{Quality Control}

To ensure the reliability and consistency of SpeechR, we implement a multi-stage quality control pipeline designed to verify the 1) textual accuracy, 2) text-audio alignment, and 3) audio human-likeness.

(1) To reduce annotation errors and improve the reliability of acoustic labels, we conduct secondary validation using GPT-4o. For each instance, GPT-4o receives the transcription, candidate answers, and ground-truth label, and independently verifies its correctness. For acoustic annotations (stress and emotion), we generate three predictions and select the most frequent one. Conflicting cases are flagged for manual review to ensure robustness.

(2) All synthesized speech samples are verified for alignment with their corresponding transcriptions. First, we ensure the existence of audio file to avoid broken links or missing data. Then, we re-transcribe the generated audio using an ASR model and apply forced alignment techniques to detect mismatches between ASR output and the original text. This guarantees strict consistency between the generated speech and the reference transcription, a critical requirement for evaluating speech-based reasoning.

(3) Although the audio in SpeechR is synthesized, the use of the Azure Speech SDK ensures naturalness and high-quality in the generated speech.
we assessed the human-likeness of the synthesized audio by using perceptual evaluation with native listeners.
we conducted a listening test with 10 native English speakers. Each participant was asked to rate 30 randomly sampled audio on a 5-point human-likeness scale (1 = robotic and unnatural, 5 = indistinguishable from natural human speech). The resulting average score was 4.8, indicating that the generated audio in SpeechR is highly natural and suitable for evaluating spoken language understanding in LALMs.

\vspace{-2mm}
\subsection{Benchmark Statistics}
As shown in Figure~\ref{overview}, SpeechR includes 3,366 multimodal reasoning instances covering three major categories: factual, procedural, and normative reasoning. The dataset is organized into three formats: a multiple choice version, a generative version, and an acoustic feature version. Each format is designed to reflect different cognitive demands, including discrete retrieval, open-ended reasoning, and prosody-informed interpretation.

On average, each transcription contains 35 words and corresponds to an audio duration of 14 seconds, with sample lengths ranging from 2.06 to 62.10 seconds. SpeechR features 37 American English voices and 15 emotional tones, synthesized using the Azure Speech SDK to ensure expressive and diverse speech data. A detailed comparison with existing benchmarks is provided in the Appendix.
\vspace{-2mm}

\section{Experiments}


\subsection{Experimental setup}
\textbf{Models} We evaluate representative LALMs on the SpeechR benchmark, which includes a diverse range of reasoning tasks spanning factual, procedural, and normative categories. For the open-source models (LTU~\cite{gong2023listen}, GAMA~\cite{ghosh2024gama}, SALMONN~\cite{tang2023salmonn}, Qwen-Audio series~\cite{chu2023qwen,chu2024qwen2}, LLaMA-Omni~\cite{fang2024llama} and Mellow~\cite{deshmukh2025mellow}) we perform local inference using published checkpoints; for proprietary services (GPT-4o-audio~\cite{openai_chatgpt} and Gemini 1.5-Pro~\cite{reid2024gemini}), we query via their APIs under default settings. All models are evaluated under a unified input format, using the same prompts, and are executed with default hyperparameters. Inference was performed on a NVIDIA A800 GPU. 

\textbf{Evaluation Protocol}
Each SpeechR version is evaluated using a format-specific protocol.

(1) Multiple-choice version.
We use a discrete-choice evaluation, where model outputs are scanned for valid option labels (e.g., “A”, “B”) and matched to the ground-truth answer. Accuracy is the proportion of correct predictions.

(2) Generative version.
We adopt an LLM-as-a-judge framework with GPT-4o. To reduce evaluation bias, model outputs are passed without rephrasing or post-processing. The judge receives only the question, model prediction, and reference answer, and scores responses blindly without access to model identity, using the following rubric:
\begin{itemize}
\item \textbf{Final Correctness} (0 or 1): Binary score indicating whether the answer matches the reference.
\item \textbf{Logical Relevance} (1 to 5, integer): Whether the answer logically follows from the question.
\item \textbf{CoT Coherence} (1 to 5, integer): Whether the reasoning is internally consistent and well-structured.
\end{itemize}
All scores are assigned as discrete integers. Higher scores on Logical Relevance and CoT Coherence reflect stronger reasoning performance. Full scoring guidelines and examples are provided in Appendix.

(3) Acoustic-feature version.
This version follows the same discrete-choice protocol as the multiple-choice format to assess how prosodic and emotional variations affect classification. Accuracy remains the primary metric.

Together, these protocols support both accuracy-based classification and fine-grained evaluation of open-ended reasoning, offering a comprehensive assessment of LALMs across spoken formats.

\subsection{Main Results}

\subsubsection{Results on Multiple-Choice Version}
First, we evaluated eleven state-of-the-art LALMs on SpeechR using the discrete-choice protocol, with tasks ranging from binary to five-way classification. Each model was prompted to produce either a binary decision or select the correct option, with accuracy as the primary evaluation metric. 
\paragraph{Discussion}
As shown in Table~\ref{tab:core-full}, LALM performance on the SpeechR multiple-choice benchmark exhibits substantial variation across reasoning categories, offering several insights into their current strengths and limitations.

First, advanced proprietary models such as GPT-4o-audio-preview and Gemini-1.5-Pro consistently outperform others, especially on factual and procedural tasks. This suggests that large-scale pretraining, combined with strong audio-language integration, remains effective for improving reasoning under speech input.

However, notable challenges remain. Even the best-performing models exhibit degraded accuracy on tasks requiring nuanced social inference, such as Moral Judgment. This suggests that LALMs still struggle to model the pragmatic subtleties and context dependencies inherent in conversational normative reasoning.

Importantly, we observe a marked performance drop when comparing model results on SpeechR to their text-only counterparts. For instance, tasks like GSM8K and BoolQ routinely exceed 85\% accuracy in models such as GPT-4o and Qwen-2.5 under standard text-based evaluation. Yet, when converted into spoken form with cleaned reasoning prompts, accuracy drops significantly—even for the most capable LALMs. This highlights that speech reasoning involves more than transcribing input: it requires robust multimodal alignment and context integration across acoustic and linguistic channels.

Together, these findings suggest that despite strong language backbones, LALMs still struggle with reasoning under speech input. Future research should focus on enhancing spoken context understanding, diversifying training data, and improving fusion of acoustic and semantic cues.

\begin{table}[t]
\centering
\small 
\setlength{\tabcolsep}{3pt} 
\renewcommand{\arraystretch}{1.05}  

\begin{tabular}{lccc|cc}
\toprule
\textbf{Model} & \textbf{FC} & \textbf{LR} & \textbf{Coh} & \textbf{FC} & \textbf{LR} \\
 & \multicolumn{3}{c|}{Procedural} & \multicolumn{2}{c}{Normative} \\
\midrule
Mellow & 16.96 & 2.42 & 1.64 & 30.92 & 2.59 \\
SALMONN & 12.50 & 1.90 & 1.33 & 34.75 & 3.03 \\
Qwen-Audio-Chat & 3.72 & 1.67 & 1.53 & 34.01 & 2.93 \\
Qwen2-Audio-7B & 9.52 & 1.50 & 1.00 & 31.25 & 2.82 \\
Qwen2-Audio-Instruct & 25.00 & 3.50 & 2.16 & 38.91 & 3.46 \\
LLaMA-Omni & 17.86 & 3.14 & 2.44 & 33.26 & 3.24 \\
GPT-4o-preview & \textbf{89.43} & \textbf{4.80} & \textbf{4.71} & 50.21 & \textbf{3.59} \\
Gemini-1.5-Pro & 83.04 & 4.49 & 4.47 & \textbf{51.92} & 3.58 \\
\bottomrule
\end{tabular}

\caption{Performance of LALMs on the SpeechR generative version. 
FC = final correctness, LR = logical relevance, Coh = coherence.}
\label{tab:core-gen}
\vspace{-5mm}
\end{table}

\subsubsection{Results on Generative Version}
We evaluate eight LALMs on the SpeechR generative version to assess open-ended reasoning.
Three models from the discrete-choice evaluation were excluded due to a lack of observable chain-of-thought generation during preliminary testing. The llm-as-a-judge protocol is tailored to each task category. For moral questions, we record only final correctness and logical relevance. For procedural CoT prompts, which require structured multi-step inference, we apply all three metrics: final correctness, logical relevance, and CoT coherence. This setup captures both the validity of the answer and the internal logical structure.

\paragraph{Discussion}
As shown in Table~\ref{tab:core-gen}, most LALMs struggle in the generative setting, especially on normative tasks such as moral judgment. These tasks require open-ended reasoning, sensitivity to social norms, and coherence across two-speaker interactions, all of which remain challenging for current models. In contrast, GPT-4o-audio and Gemini-1.5-Pro perform strongly on procedural items, demonstrating clear multi-step reasoning and logical structure. Their success can be attributed to large-scale pretraining and instruction tuning with chain-of-thought data. Qwen2-Audio-Instruct also shows competitive results, benefiting from extensive instruction tuning. However, other models lack consistent reasoning chains and often fail to maintain logical coherence. These findings suggest that improving LALMs' generative reasoning, particularly in socially complex domains, requires further tuning with rich, structured reasoning examples and more diverse discourse formats.

\begin{table}[t]
\vspace*{-2mm}  
\centering
\small  
\setlength{\tabcolsep}{3pt}  
\renewcommand{\arraystretch}{1.05}  

\begin{tabular}{lcccc}
\toprule
\textbf{Model} & \textbf{Base} & \textbf{Stress} & \textbf{Emotion} & \textbf{Both} \\
\midrule
LTU & 32.93 & 32.34 & 31.74 & 33.23 \\
GAMA & 33.83 & 35.93 & 35.33 & 32.63 \\
GAMA-IT & 20.36 & 18.26 & 17.07 & 19.46 \\
Mellow & 23.95 & 23.95 & 24.25 & 26.65 \\
SALMONN & 34.43 & 33.83 & 33.53 & 33.83 \\
Qwen-Audio-Chat & 33.53 & 33.23 & 33.53 & 31.74 \\
Qwen2-Audio-7B & 9.88 & 10.18 & 10.18 & 9.58 \\
Qwen2-Audio-Instruct & 34.13 & 33.83 & 32.33 & 32.93 \\
LLaMA-Omni & 38.32 & 37.72 & 36.53 & 35.93 \\
GPT-4o-preview & 57.78 & 55.39 & 60.78 & 55.09 \\
Gemini-1.5-Pro & \textbf{64.67} & \textbf{64.37} & \textbf{64.97} & \textbf{65.87} \\
\bottomrule
\end{tabular}

\caption{Accuracy (\%) of LALMs on the SpeechR acoustic-feature version.}
\label{tab:core-acoustic}
\vspace*{-4mm}  
\end{table}

\subsubsection{Results on Acoustic-Feature Version}
We evaluate the SpeechR acoustic-feature version under four conditions: original, stress-modified, emotion-modified, and combined audio. All evaluations follow the discrete-choice protocol with accuracy as the primary metric. Results in Table~\ref{tab:core-acoustic} show how prosodic stress and emotional tone individually and jointly affect the reasoning performance of LALMs.

\paragraph{Discussion}
The acoustic-feature version is designed to examine how prosodic stress and emotional tone influence spoken reasoning. Evaluation results reveal that LALMs respond differently to these acoustic variations, offering insight into how speech expressiveness affects model behavior. Instruction-tuned models such as Qwen2-Audio-Instruct and GAMA show minor performance shifts when acoustic features are introduced, suggesting that while these models can process surface-level cues, their reasoning behavior remains largely invariant to expressive signals.

Interestingly, Mellow demonstrates a moderate performance increase under the Emotion and Both conditions. This may be attributed to its dual-encoder architecture, which encourages sensitivity to fine-grained acoustic cues. Such a result indicates that certain model architectures may be more responsive to expressive speech and able to integrate prosodic signals into the reasoning process.

Large-scale models like GPT-4o-audio and Gemini-1.5-Pro maintain strong overall performance, but their accuracy drops slightly under stress-enhanced speech. This suggests that while these models are highly capable in general, their reasoning may still be affected by subtle changes in speech delivery, particularly when emphasis patterns are altered.

These findings underscore the importance of modeling not only the content but also the expressive form of speech. By isolating specific prosodic factors, the acoustic-feature version enables deeper investigation into how acoustic expressiveness shapes reasoning outcomes in LALMs, offering a new perspective on multimodal language understanding.

\section{Conclusion}
We present SpeechR, a benchmark designed to evaluate spoken-language reasoning in large audio-language models. It covers three reasoning types: factual, procedural, and normative, and includes three evaluation formats: multiple-choice, generative, and acoustic-feature versions.

Results show that while some models perform well on factual tasks, most still struggle with procedural and normative reasoning. This highlights the need for better instruction tuning and the integration of acoustic cues into reasoning processes.

SpeechR provides a foundation for future research aimed at building more capable and context-aware audio-language models. We hope SpeechR can guide the development of safer, more socially-aware audio-language systems for applications in education, accessibility, and digital assistants.


\bibliography{aaai2026}




\section{Details of SpeechR Dataset}
\label{sec:dataset_details}

\subsection{Readability Enhancement}

We apply a set of controlled rewriting rules to ensure that the speech is syntactically well-formed, semantically faithful, and suitable for high-quality audio synthesis. Specifically, we adopt the following criteria:

\begin{itemize}
\item The text length is constrained to fall within 7 to 150 words.
\item All emojis and special characters (e.g., ``:)'' or ``\#'') are removed.
\item Abbreviations and incomplete words are expanded into their full forms (e.g., ``tkts'' becomes ``tickets'', ``comp'' becomes ``competition'').
\item URL links are simplified to include only the domain name.
\item Appropriate punctuation and sentence segmentation are added to improve clarity and naturalness when read aloud.
\item No new content is introduced beyond what is present in the original input.
\end{itemize}

\subsection{Interaction Enrichment}

To simulate natural conversational dynamics, we enhance the interactivity of the original text-based datasets using two strategies: restructuring the data format and modifying pronoun perspectives.

\paragraph{Restructuring the Data Format}

Some source datasets do not follow a dialogue or question-answer format. For these, we convert each instance into a unified conversational structure.

For example, in the DailyDilemmas dataset, each instance includes a moral scenario and the outcomes of different actions (e.g., choosing to act or not). We construct the \textbf{Person A} utterance by combining the scenario and a corresponding action-related question, and generate the \textbf{Person B} response by combining the selected action and its consequence. This transformation results in a coherent two-turn dialogue.

\paragraph{Modifying Pronoun Perspectives}

To make the interaction feel more natural, we insert personal pronouns into the conversation. Specifically:
\begin{itemize}
\item The \textbf{Person A} prompt is rephrased to include ``you'' (e.g., ``What should one do in this situation?'' $\rightarrow$ ``What should you do in this situation?'').
\item The \textbf{Person B} response is rephrased to include ``I'' (e.g., ``One should avoid this.'' $\rightarrow$ ``I would avoid this.'').
\end{itemize}

These adjustments ensure that the resulting dialogue better mirrors speaker-centric, conversational speech patterns.

\subsection{Annotation of Acoustic Features}

To enrich SpeechR with acoustic cues relevant to reasoning, we annotate each instance with two types of features: stress and emotion. Annotations are generated using GPT-4o under structured prompt templates, as described below.

\textbf{Stress Annotation.}  
We provide GPT-4o with the transcribed text and the ground-truth answer, asking it to identify the keyword or phrase within the transcription that most strongly determines the correct answer. The model is explicitly instructed to select the word or phrase that should be emphasized if the sentence were spoken aloud.

\textbf{Emotion Annotation.}  
For emotion annotation, GPT-4o is given the transcribed text, the ground-truth answer, and a predefined list of emotions. It is instructed to select the emotion most appropriate for reading the text aloud, considering both the tone of the content and the underlying reasoning intent.

These annotations are used to assess whether LALMs can benefit from or are sensitive to prosodic and emotional features during reasoning tasks.

\subsection{Comparison with Existing Benchmarks}

We identify four major limitations in existing audio datasets:

(1) Most are designed for low-level audio understanding tasks such as event classification or speech recognition, rather than high-level reasoning. For example, MELD and IEMOCAP primarily target emotion recognition, while VoxCeleb focus on speaker identification.

(2) Many datasets fail to simulate realistic dialogue reasoning scenarios, limiting their applicability to natural conversational inference. For instance, MMAU provides an audio dialogue and asks the LALM to identify the roles of the speakers. However, such reasoning is relatively straightforward and lacks inferential depth. In most human or human-machine conversations, speaker roles are either explicitly stated or easily inferred from surface cues. Therefore, it fails to capture the complexity required for evaluating real-world reasoning.

(3) Existing datasets often lack diversity in reasoning types. For example, temporal reasoning and content-based reasoning benchmarks focus narrowly on specific inference categories, limiting their ability to provide a comprehensive assessment of the reasoning capabilities of LALMs.

(4) Even among reasoning-oriented datasets, the complexity of reasoning remains constrained. For instance, although MMAU introduces dialogue-like audio data, its short average audio length (approximately 10 seconds) significantly limits the depth of reasoning it can support.

In contrast, our SpeechR benchmark addresses these limitations by offering a broader range of reasoning tasks, spanning factual, procedural, and normative dimensions. It incorporates more realistic and context-rich dialogue scenarios, simulating natural conversational settings. Furthermore, SpeechR supports diverse output formats, including both multiple-choice and generative responses, and emphasizes clearer, more complex reasoning chains that reflect the multi-step inference required in real-world audio interactions. These features enable a more comprehensive and challenging evaluation of LALMs' reasoning capabilities.

\section{Model Descriptions}
\label{sec:models}

\paragraph{LTU}
LTU is a multimodal large language model designed for general audio understanding. Trained on the OpenAQA-5M dataset, it demonstrates strong performance in audio classification and captioning tasks.

\paragraph{GAMA}
GAMA is a general-purpose large audio-language model that integrates multiple types of audio representations. Fine-tuned with the CompA-R dataset, it enhances complex audio reasoning abilities, outperforming other LALMs in diverse audio understanding tasks.

\paragraph{GAMA-IT}
An instruction-tuned variant of GAMA, GAMA-IT is designed to improve performance in open-ended audio question-answering tasks requiring complex reasoning. It leverages instruction tuning to enhance its reasoning capabilities.

\paragraph{Mellow}
Mellow is a compact 167M parameter audio-language model optimized for reasoning tasks. It takes in two audio inputs and a text prompt, producing free-form text outputs. Despite its small size, Mellow achieves competitive performance with significantly fewer resources.

\paragraph{SALMONN}
SALMONN is a large language model enabling speech, audio events, and music inputs. It supports various tasks, including automatic speech recognition, emotion recognition, and audio question-answering.

\paragraph{Qwen-Audio-Chat}
Qwen-Audio-Chat is a multimodal model that accepts diverse audio inputs and text. It is designed for tasks such as speech recognition and audio-text understanding, emphasizing instruction-following capabilities.

\paragraph{Qwen2-audio-7B}
An updated large-scale audio-language model, Qwen2-Audio-7B is capable of handling various audio signals and performing audio analysis or direct textual responses.

\paragraph{Qwen2-audio-Instruct}
This is an instruction-tuned version of Qwen2-Audio-7B, enhancing the model's ability to follow prompts and perform complex reasoning tasks.

\paragraph{LLama-Omni}
LLaMA-Omni is a speech-language model supporting low-latency, high-quality speech interactions. It can generate both text and speech responses directly from speech instructions with extremely low latency.

\paragraph{GPT-4o-audio-preview}
OpenAI's GPT-4o-Audio-Preview is a multimodal model integrating real-time audio, vision, and text processing capabilities. It enables natural speech interactions and multilingual translation, allowing users to talk to ChatGPT with real-time responses and interruptions.

\paragraph{Gemini-1.5-pro}
Gemini-1.5-pro is an advanced multimodal model supporting text, code, image, audio, and video inputs. It is designed for high-efficiency reasoning and generation tasks, with capabilities in understanding and interacting with various data modalities.

\section{Source Datasets}
\label{sec:datasets}

In this section, we introduce the set of datasets utilized in the construction of the SpeechR benchmark.

\paragraph{ReClor}  
ReClor is a reading comprehension dataset consisting of logical reasoning questions derived from standardized graduate-level entrance exams. Each instance includes a passage, a question, and multiple-choice answers, with a strong focus on deductive reasoning.

\paragraph{BoolQ}  
BoolQ is a yes/no question-answering dataset where each question is paired with a short supporting passage. The questions are naturally occurring and require understanding of factual content from the given context.

\paragraph{CommonsenseQA}  
CommonsenseQA is a multiple-choice question answering dataset that targets commonsense reasoning. It is built on ConceptNet relations and presents challenging examples that often require reasoning beyond surface-level word matching.

\paragraph{RiddleSense}  
RiddleSense is a dataset designed to evaluate lateral thinking and creative commonsense reasoning. It contains multiple-choice riddles, where the correct answer requires both semantic understanding and reasoning through implicit associations.

\paragraph{GSM8K}  
GSM8K is a math word problem dataset designed to assess grade-school level arithmetic reasoning. It includes detailed step-by-step chain-of-thought annotations, making it a standard benchmark for evaluating procedural reasoning in CoT settings.

\paragraph{ReveAL-CoT}  
ReveAL-CoT is a scientific reasoning dataset featuring multi-step inference questions across physics, biology, and other science domains. Each question is annotated with chain-of-thought explanations to support structured procedural reasoning.

\paragraph{ETHICS}  
ETHICS is a benchmark for moral reasoning that includes scenarios requiring judgments about the ethical permissibility of actions. The dataset covers diverse moral contexts such as fairness, harm, and loyalty.

\paragraph{DailyDilemmas}  
DailyDilemmas contains short narratives that describe everyday situations involving social or ethical decision-making. Each instance asks the model to evaluate the appropriateness or morality of an individual's behavior.

\paragraph{SMS Spam Collection}  
This dataset contains a collection of labeled SMS messages, with each message categorized as spam or ham (not spam). It is widely used for binary classification tasks involving deception or malicious intent detection.

\paragraph{Enron Email}  
The Enron Email dataset comprises real-world corporate email communications, with a subset labeled for spam detection. It supports studies in behavioral and normative analysis, especially in identifying unethical or misleading content.

\section{Prompt Templates}
\label{sec:prompts}

\begin{figure*}[ht]
\centering
  \includegraphics[width=1.0\textwidth]{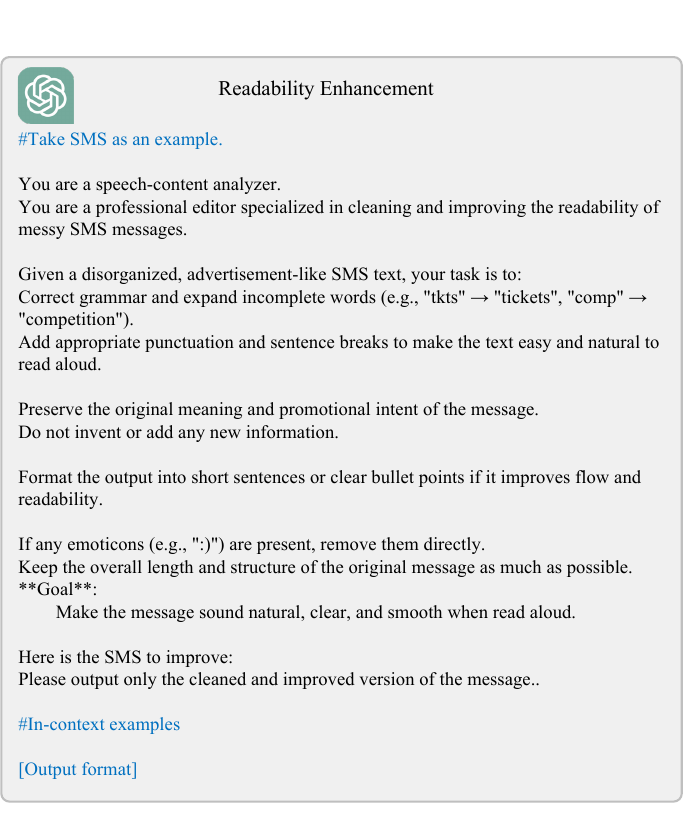}
  \caption{Prompt template for dataset readability enhancement.}
    \label{fig:read}
\end{figure*}

\begin{figure*}[ht]
\centering
  \includegraphics[width=1.0\textwidth]{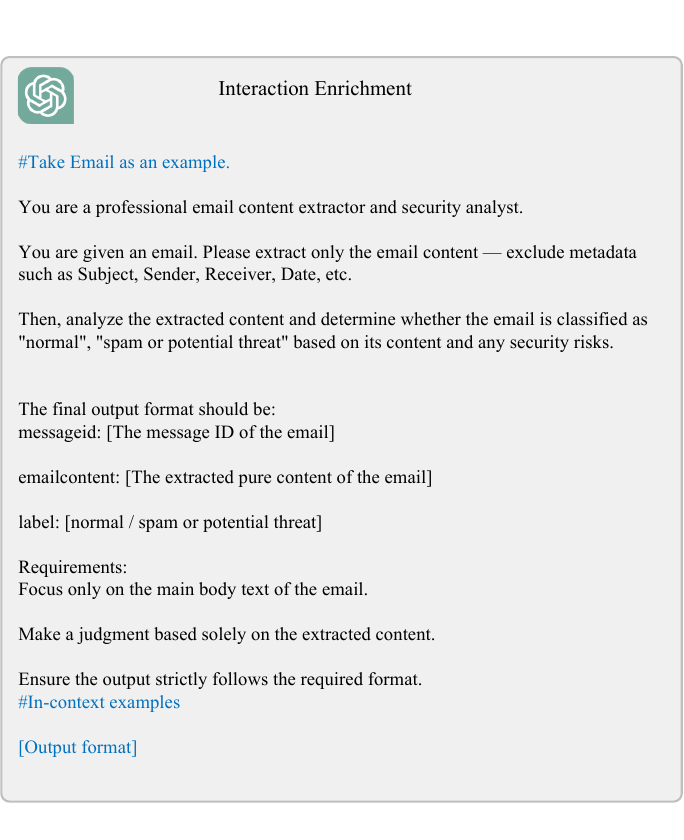}
  \caption{Prompt template for dataset interaction enrichment.}
    \label{fig:inter}
\end{figure*}

\begin{figure*}[ht]
\centering
  \includegraphics[width=1.0\textwidth]{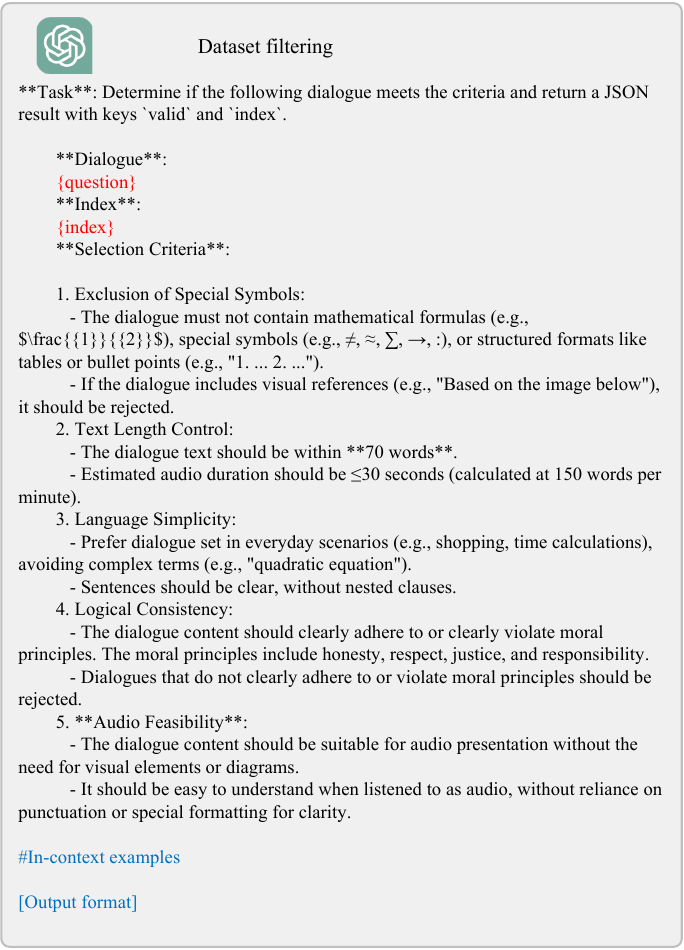}
  \caption{Prompt template for dataset filtering.}
    \label{fig:filter}
\end{figure*}

In this section, we present the prompt templates used throughout various stages of our dataset construction and evaluation pipeline. Figures~\ref{fig:read},~\ref{fig:inter}, and~\ref{fig:filter} illustrate three prompt designs employed during the generation of the SpeechR dataset. Specifically, Figure~\ref{fig:read} shows the prompt used to enhance readability and linguistic fluency of raw samples, Figure~\ref{fig:inter} demonstrates the interaction-oriented prompt that encourages more engaging and context-aware formulations, and Figure~\ref{fig:filter} presents the filtering prompt used for quality control, enabling the exclusion of incoherent or irrelevant data.

\begin{figure*}[ht]
\centering
  \includegraphics[width=1.0\textwidth]{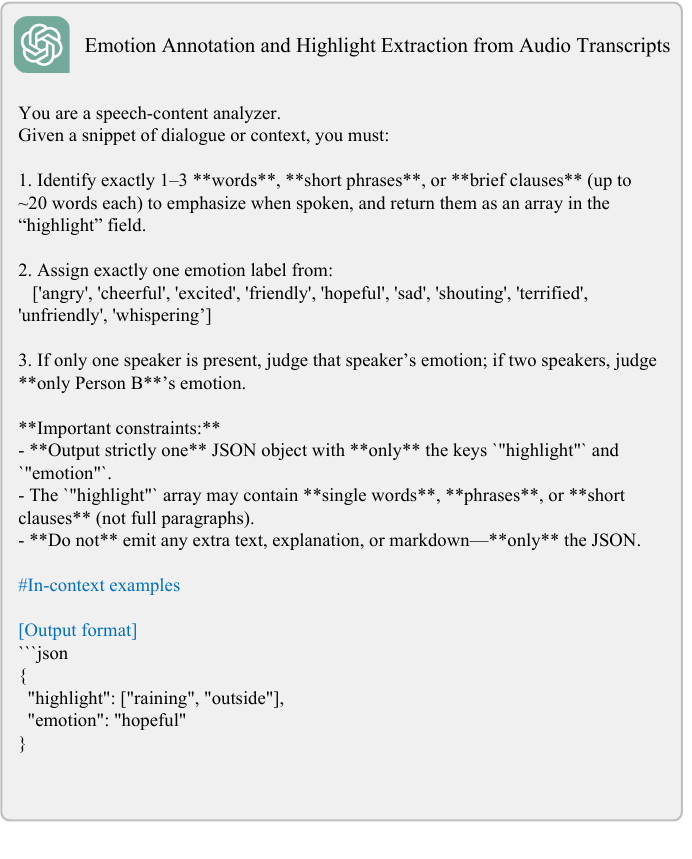}
  \caption{Prompt template for emotion annotation and highlight extraction from audio transcripts for SpeechR mini version.}
    \label{fig:llm_emotion}
\end{figure*}

\begin{figure*}[ht]
\centering
  \includegraphics[width=1.0\textwidth]{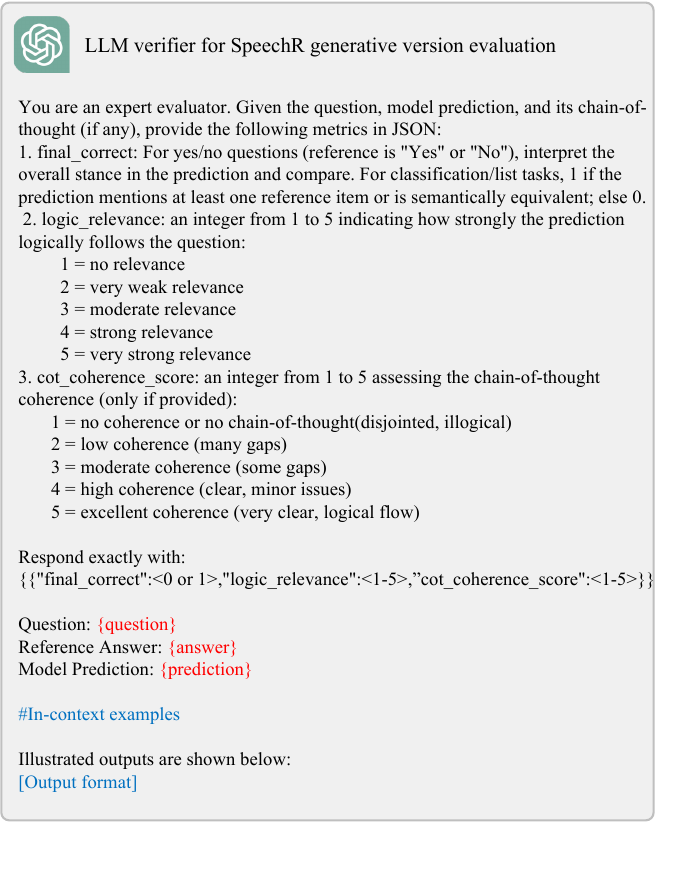}
  \caption{Prompt template for LLM-Based evaluation of the SpeechR generative version.}
    \label{fig:llm_eval}
\end{figure*}

In addition, Figures~\ref{fig:llm_emotion} and~\ref{fig:llm_eval} display prompt templates used in the evaluation phase. Figure~\ref{fig:llm_emotion} is designed for emotion annotation and highlight word extraction from audio transcripts, and is applied specifically to the mini version of SpeechR. Figure~\ref{fig:llm_eval} illustrates the prompt format adopted for LLM-as-a-judge evaluation of the generative version, guiding the model to assess response correctness, logical relevance, and reasoning coherence.

\section{Qualitative Analysis}
\label{sec:qualitative}

As shown in Table~\ref{tab:quali}, we present qualitative results from different LALMs across the three reasoning categories in SpeechR: factual reasoning, illustrated with creative puzzles; procedural reasoning, represented by mathematical problems; and normative reasoning, which highlights the models' ability to generate inferences in dialogue-based scenarios.

\begin{table*}[t]
\centering
\begin{tabular*}{\linewidth}{l}
\toprule
\textbf{(Factual Reasoning)}\\
There was a green house in the green house was a white house in that white house was a red house \\in that red house where eggs. What is it?\\
A: salt \quad B: jar \quad C: the park \quad D: supermarket \quad E: watermelon\\
\textbf{Ground Truth:} E\\
\midrule
\textbf{LTU}: \textcolor{blue}{The best option is A. salt.}\\
\textbf{GAMA}: \textcolor{blue}{The best option is D. supermarket. The audio mentions 'the park' and 'watermelon',} \\
\textcolor{blue}{but not 'supermarket'.}\\
\textbf{GAMA-IT}: \textcolor{blue}{The audio does not provide enough information to accurately determine the location.}\\
\textbf{Mellow}: \textcolor{blue}{a particular way is best said about the best option...}\\
\textbf{SALMONN}: E. watermelon\\
\textbf{Qwen-Audio-Chat}: \textcolor{blue}{The best option is C. the park.}\\
\textbf{Qwen-Audio-7B}: \textcolor{blue}{what is it?}\\
\textbf{Qwen-Audio-Instruct}: \textcolor{blue}{It is a riddle that describes a series of houses based on their colors.}\\
\textcolor{blue}{The answer to the riddle is Egg}\\
\textbf{LLaMA-Omni}: \textcolor{blue}{The correct answer is B. jar}\\
\textbf{GPT-4o}: The riddle describes a watermelon. The green house refers to the green outer rind, \\the white house refers to the white inner layer, the red house refers to the red flesh, and the eggs \\refer to the seeds inside. Therefore, the best option is E. watermelon.\\
\textbf{Gemini-1.5-Pro}: The riddle describes a watermelon. Therefore the answer is \textbf{E. watermelon}.\\
\midrule
\textbf{(Procedural Reasoning)}\\
John takes care of 10 dogs. Each dog takes .5 hours a day to walk and take care of their business. \\How many hours a week does he spend taking care of dogs?\\
A: 65 \quad B: 35 \quad C: 32.5 \quad D: 30\\
\textbf{Ground Truth:} B\\
\midrule
\textbf{LTU}: B.35\\
\textbf{GAMA}: \textcolor{blue}{A.35}\\
\textbf{GAMA-IT}: \textcolor{blue}{The correct option is D.30}\\
\textbf{Mellow}: \textcolor{blue}{A}\\
\textbf{SALMONN}: B.35\\
\textbf{Qwen-Audio-Chat}: \textcolor{blue}{The answer is A.65.}\\
\textbf{Qwen-Audio-7B}: \textcolor{blue}{how many hours a week does he spend taking care of dogs?}\\
\textbf{Qwen-Audio-Instruct}: \textcolor{blue}{John spends 7 hours a week taking care of dogs.}\\
\textbf{LLaMA-Omni}: \textcolor{blue}{John spends fifty hours a week taking care of dogs.}\\
\textbf{GPT-4o}: B.35\\
\textbf{Gemini-1.5-Pro}: B\\
\midrule
\textbf{(Normative Reasoning)}\\
\textbf{Person A}: You're a project manager in a marketing firm. You notice a colleague using company \\resources for a personal project during office hours, undermining the company's policy. \\Should you take initiative and report this to your superior? \\
\textbf{Person B}: Yes. I report the colleague to my superior. The colleague is reprimanded and harbors \\resentment towards me, creating a hostile work environment. This affects team morale \\and productivity, leading to a decline in the quality of work.\\
\textbf{Question}: What moral values are reflected in this response?\\
\textbf{Ground Truth:} ``respect for rules'', ``accountability'', ``responsibility'', ``honesty'', ``trust'', ``integrity'', ``fairness''\\
\midrule
\textbf{Mellow}: \textcolor{blue}{The response from the other woman is a brief, soft utterance, likely a comment...}\\
\textbf{SALMONN}: The speaker values taking initiative and reporting unethical behavior to their superior.\\
\textbf{Qwen-Audio-Chat}: The second speaker mentions that they value honesty and integrity \\in the workplace.\\
\textbf{Qwen-Audio-7B}: \textcolor{blue}{[Music]}\\
\textbf{Qwen-Audio-Instruct}: \textcolor{blue}{I cannot provide a definitive answer as it depends on various factors}\\
\textcolor{blue}{such as company policies, cultural norms, and individual personalities.} \\
\textbf{LLaMA-Omni}: \textcolor{blue}{I would take initiative and report this to my superior.}\\
\textbf{GPT-4o}: The speaker values adherence to company policies and is willing to take action \\to uphold them, even if it may lead to personal or team conflict\\
\textbf{Gemini-1.5-Pro}: The speaker prioritizes upholding company policy and addressing misconduct, \\even at personal cost.\\
\bottomrule
\end{tabular*}
\caption{Examples of responses from LALMs across different reasoning categories. \textcolor{blue}{Blue text} indicates abnormal or incorrect responses.}
\label{tab:quali}
\end{table*}

\section{Reflections and Future Directions}
\label{sec:limitations}

SpeechR is designed as an initial step toward evaluating reasoning in speech-based interactions. While it covers a diverse range of reasoning tasks and introduces controlled acoustic variations, further extensions may broaden its scope in the following directions:

\paragraph{Speech Variability}
All speech in SpeechR is consistently synthesized using standardized settings. Expanding to include more varied prosody, speaking styles, and spontaneous speech could offer richer insights into real-world model performance.

\paragraph{Linguistic and Cultural Coverage}
Current data construction focuses on English with general social contexts. Exploring additional languages and sociocultural scenarios could enable broader applicability across multilingual and multicultural settings.

\paragraph{Interaction Dynamics}
The current benchmark emphasizes static single-turn prompts. Incorporating multi-turn dialogue and speaker dynamics could allow future benchmarks to capture more interactive aspects of speech-based reasoning.

\end{document}